\documentclass[10pt,journal,compsoc]{IEEEtran}

\usepackage{textcomp}
\usepackage{graphics} 
\usepackage{graphicx, caption, subcaption}
\usepackage{color}
\usepackage{bm}
\usepackage{float}
\usepackage{booktabs}
\usepackage{mathptmx}
\usepackage{amsmath}

\ifCLASSOPTIONcompsoc
  \usepackage[nocompress]{cite}
\else
  \usepackage{cite}
\fi

\begin{document}

\title{Attribute-conditioned Layout GAN \\ for Automatic Graphic Design}

\author{Jianan~Li, 
        Jimei~Yang, 
        Jianming~Zhang, 
        Chang~Liu, 
        Christina~Wang, 
        and~Tingfa~Xu
\IEEEcompsocitemizethanks{\IEEEcompsocthanksitem J. Li and T. Xu are with Beijing Institute of Technology, 5 South Zhongguancun Street, Haidian District, Beijing 100081, China. E-mail: \{20090964, xutingfa\}@bit.edu.cn.
\IEEEcompsocthanksitem J. Yang and J. Zhang and C. Liu and C. Wang are with Adobe Inc., 345 Park Ave, San Jose, CA 95110, USA. E-mail: \{jimyang, jianmzha, cliu, cwang\}@adobe.com}
}

\IEEEtitleabstractindextext{
\begin{abstract}
Modeling layout is an important first step for graphic design. Recently, methods for generating graphic layouts have progressed, particularly with Generative Adversarial Networks (GANs). However, the problem of specifying the locations and sizes of design elements usually involves constraints with respect to element attributes, such as area, aspect ratio and reading-order. Automating attribute conditional graphic layouts remains a complex and unsolved problem. In this paper, we introduce Attribute-conditioned Layout GAN to incorporate the attributes of design elements for graphic layout generation by forcing both the generator and the discriminator to meet attribute conditions. Due to the complexity of graphic designs, we further propose an element dropout method to make the discriminator look at partial lists of elements and learn their local patterns. In addition, we introduce various loss designs following different design principles for layout optimization. We demonstrate that the proposed method can synthesize graphic layouts conditioned on different element attributes. It can also adjust well-designed layouts to new sizes while retaining elements' original reading-orders. The effectiveness of our method is validated through a user study.
\end{abstract}

}

\maketitle

\IEEEdisplaynontitleabstractindextext

\IEEEpeerreviewmaketitle

\IEEEraisesectionheading{\section{Introduction}\label{sec:introduction}}
\IEEEPARstart{G}{raphic} design is very important in creating multimedia. It has a wide range of applications in areas including advertisement, books and websites, among many others. Good designs that are visually pleasing while clearly delivering information, highly rely on the expertise of experienced designers. Designers often need to create the same design in different sizes or retarget existing designs to new sizes to fit in various display resolutions, for example, from websites to mobile phone screens, which requires a great deal of human labor and intelligence. Automation tools for creating designs in desired sizes would assist designers and reduce design cycle time significantly.

The first step to render real designs is to generate graphic layouts by specifying locations and sizes of design elements. Conventional machine learning techniques, such as Generative Adversarial Networks (GANs)~\cite{goodfellow2014generative,karras2017progressive} have been used to generate natural-looking digital images in pixel space~\cite{radford2015unsupervised,zhang2017stackgan,brock2018large,karras2018style}. However, these methods are not suitable for synthesizing graphic designs which are not pixels but layouts of editable design elements. 
Recently, LayoutGAN~\cite{li2018layoutgan} has been proposed to directly generate geometric parameters of graphic elements instead of image pixels from random variables. It demonstrates the feasibility of generating parameterized graphic layouts with neural networks, but it is incapable of doing practical tasks such as automatic graphic design and retargeting when specific user data is provided. Such user data includes specific design elements and their corresponding content-based attributes. These attributes introduce additional spatial constraints to be obeyed when generating the layout.
In this paper, we introduce three important attributes: 1) \emph{ Expected area} of elements with respect to the page size needs to be preserved. Distorted sizes will affect either the readability or the aesthetics. 2) \emph{Aspect ratio} of specific elements are supposed to be retained, e.g. logos. 3) \emph{Reading-order} or spatial distribution of elements should be preserved. For example, logos often appear on top, value propositions (headlines or images) in the middle and call-to-actions on the bottom for skyscraper banner ads. How to incorporate these attributes into automatic layout generation remains an open problem.

\begin{figure}
	\centering
	\includegraphics[width=1.0\columnwidth]{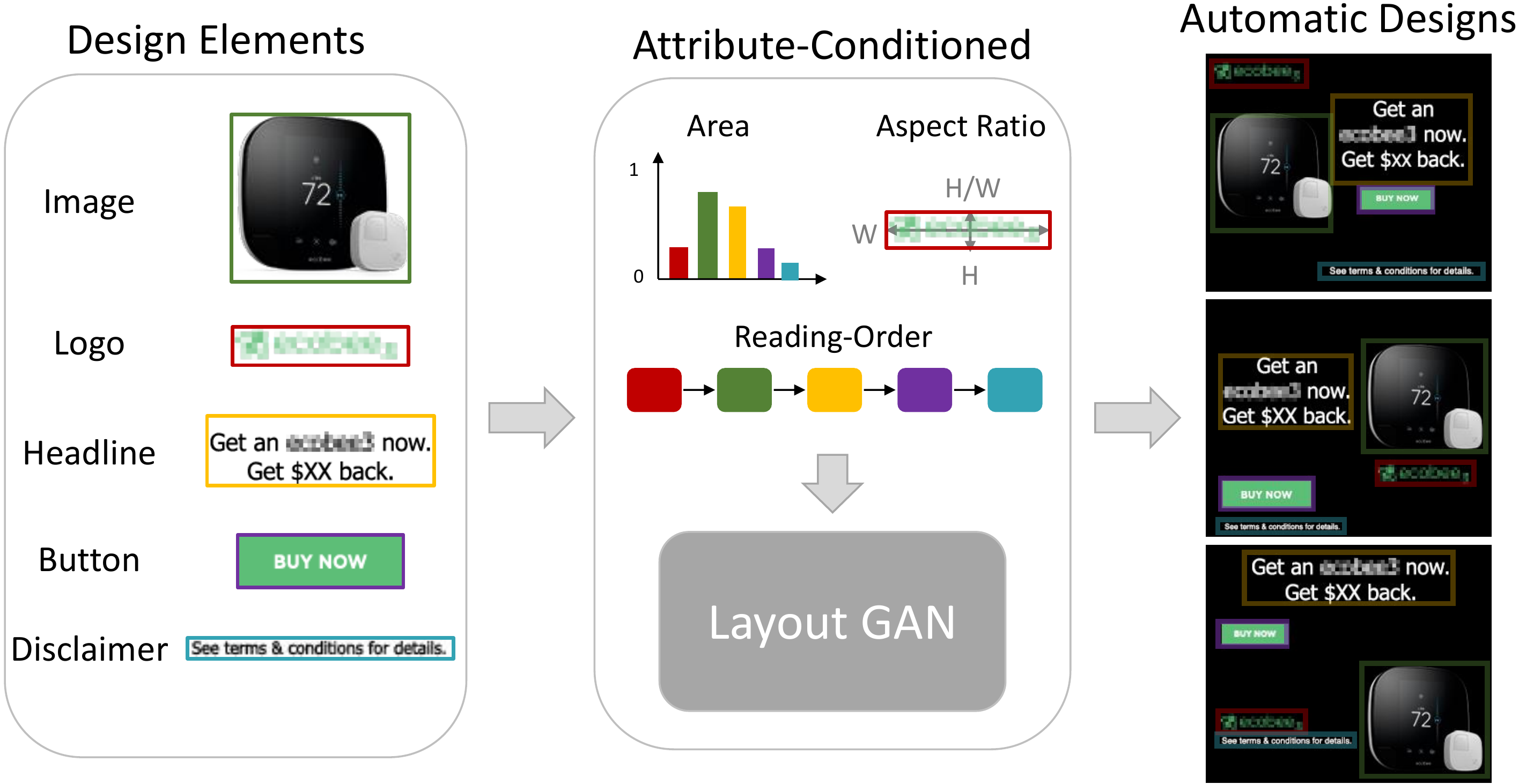}
	\caption{We present a model for automating attribute-conditioned graphic designs by incorporating different attributes such as area, aspect ratio and reading-order of given design elements.} 
	\label{fig:intro-adsdesign}
	\vspace{-2ex}
\end{figure}

In this paper, we propose an attribute-conditioned layout GAN with user data, by incorporating elements' content-based attributes as conditions for automating graphic design. As shown in Figure~\ref{fig:intro-adsdesign}, we assume a set of graphic elements along with their content-based attributes are provided as input. We introduce attribute conditions to both the generator and the discriminator of our GAN model. Specifically, we do this by directly supplying the generator with attendant attributes of each element to generate attribute conditioned layouts. Instead of directly feeding attributes to the discriminator, we introduce an additional decoder branch to reconstruct the input attributes~\cite{odena2017conditional}. 
For training the proposed GAN model, we use the wireframe rendering layer similar to LayoutGAN~\cite{li2018layoutgan} that rasterizes all parameterized elements into wireframe images. In addition to feeding all the elements to the discriminator to learn their global patterns, we also introduce a novel element dropout strategy which randomly removes some of the generated elements and feeds them to the discriminator. As there lacks global layout information delivered by a complete set of elements, the discriminator is forced to exploit local cues, such as alignment and non-overlapping of the remaining elements to better learn local patterns. 
We further propose several hand-crafted loss functions following different design principles, such as non-overlapping, alignment and preservation of reading-orders to facilitate model optimization. 

We apply our model to synthesize attribute conditional graphic layouts of different aspect ratios and retarget existing layouts to new sizes while preserving elements' reading-orders. For evaluation, we automatically generate a number of graphic designs, and compare the generated results with those produced by template-based methods, novice designers and professional designers. Experimental comparisons show that our automatic designs are better than those produced by template-based methods and novice designers, and sometimes comparable to professional designs. 

To sum up, this paper comprises the following contributions: 1) A conditional generator that synthesizes attribute conditional structured data for layout design. 2) A conditional discriminator with element dropout that exploits and learns both global and local layout patterns. 3) Several hand-crafted loss functions based on various design principles for layout optimization.

\section{Related Work}
Automatic layout generation is a major point of interest in the field of graphic design~\cite{hurst2009review,predimportance,Deka2017,kovacs2018context,ren2018charticulator}. Harrington et al.~\cite{harrington2004aesthetic} suggest an energy function to measure the aesthetics of a layout. Morcilllo et al.~\cite{morcilllo2010gaudii} create graphic designs by maximizing design quality through cost functions to meet specific design constraints. Similarly, O'Donovan et al.~\cite{o2014learning} propose an energy-based model derived from design principles to create layouts, and further develop an interactive tool~\cite{o2015designscape}. Pang et al.~\cite{pang2016directing} present a web design interaction allowing designers to direct user attention using visual flow on web designs. Swearngin et al.~\cite{swearngin2018rewire} provide an interactive system that automatically infers a vector representation of screenshots and helps designers reuse or redraw design elements. 
Li et al.~\cite{li2018layoutgan} apply generative adversarial networks to generate a set of relational graphic elements for layout design. 
Previous methods have used attribute information for design tasks with other graphic elements, such as fonts~\cite{odonovanFONT}. There are also works for 3D layout synthesis~\cite{qi2018human,yu2011make,merrell2011interactive,fisher2012example,wang2018scene,fan2017point}.

Designs are now displayed on an increasingly large variety of devices and platforms, and hence, adjusting existing graphic layouts to new sizes has attracted much attention recently. Kumar et al.~\cite{kumar2011bricolage} transfer design and content between web pages by capturing structural relationships among elements and balancing local and global concerns through optimization. Baluja et al.~\cite{baluja2006browsing} propose to retarget by segmenting a web page into coherent regions.
Cho et al.~\cite{cho2017weakly} learn an attention map which leads to a pixel-wise mapping from source to target grid for image retargeting~\cite{rubinstein2010comparative}. No previous work has applied generative adversarial networks to adjusting layouts to new sizes.

\begin{figure}[b]
	\centering    
	\begin{subfigure}{0.23\textwidth}
		\centering
		\includegraphics[width=0.9\linewidth]{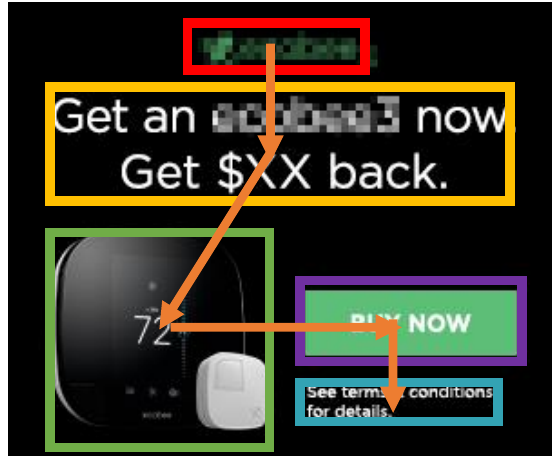}
		\caption{Professional design.}
		\label{attr-org}
	\end{subfigure}	
	\begin{subfigure}{0.23\textwidth}
		\centering
		\includegraphics[width=0.9\linewidth]{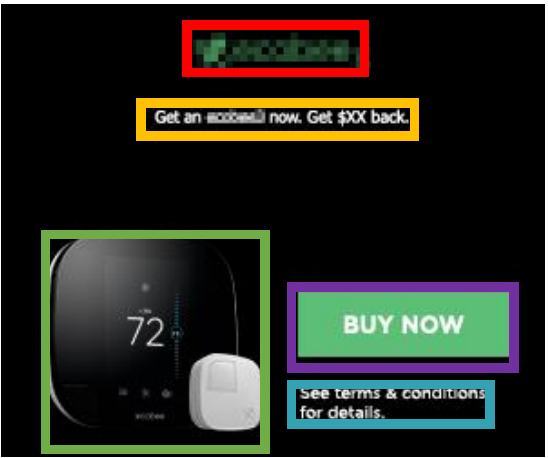}
		\caption{Variation of area.}
		\label{attr-area}
	\end{subfigure}	
	\begin{subfigure}{0.23\textwidth}
		\centering
		\includegraphics[width=0.9\linewidth]{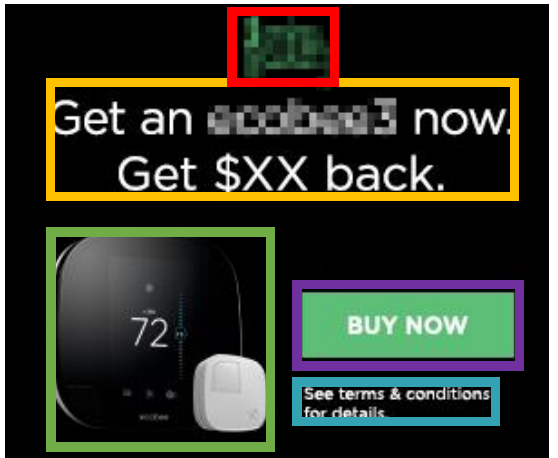}
		\caption{Variation of aspect ratio.}
		\label{attr-ratio}
	\end{subfigure}
	\begin{subfigure}{0.23\textwidth}
		\centering
		\includegraphics[width=0.9\linewidth]{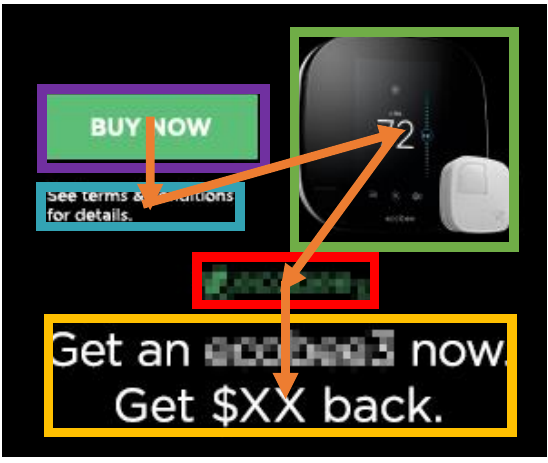}
		\caption{Variation of reading-order.}
		\label{attr-order}
	\end{subfigure}
	\caption{Attributes to be preserved for design elements.}
	\label{fig:method-attribute}	
\end{figure}

\begin{figure*}
	\centering
	\includegraphics[width=1.95\columnwidth]{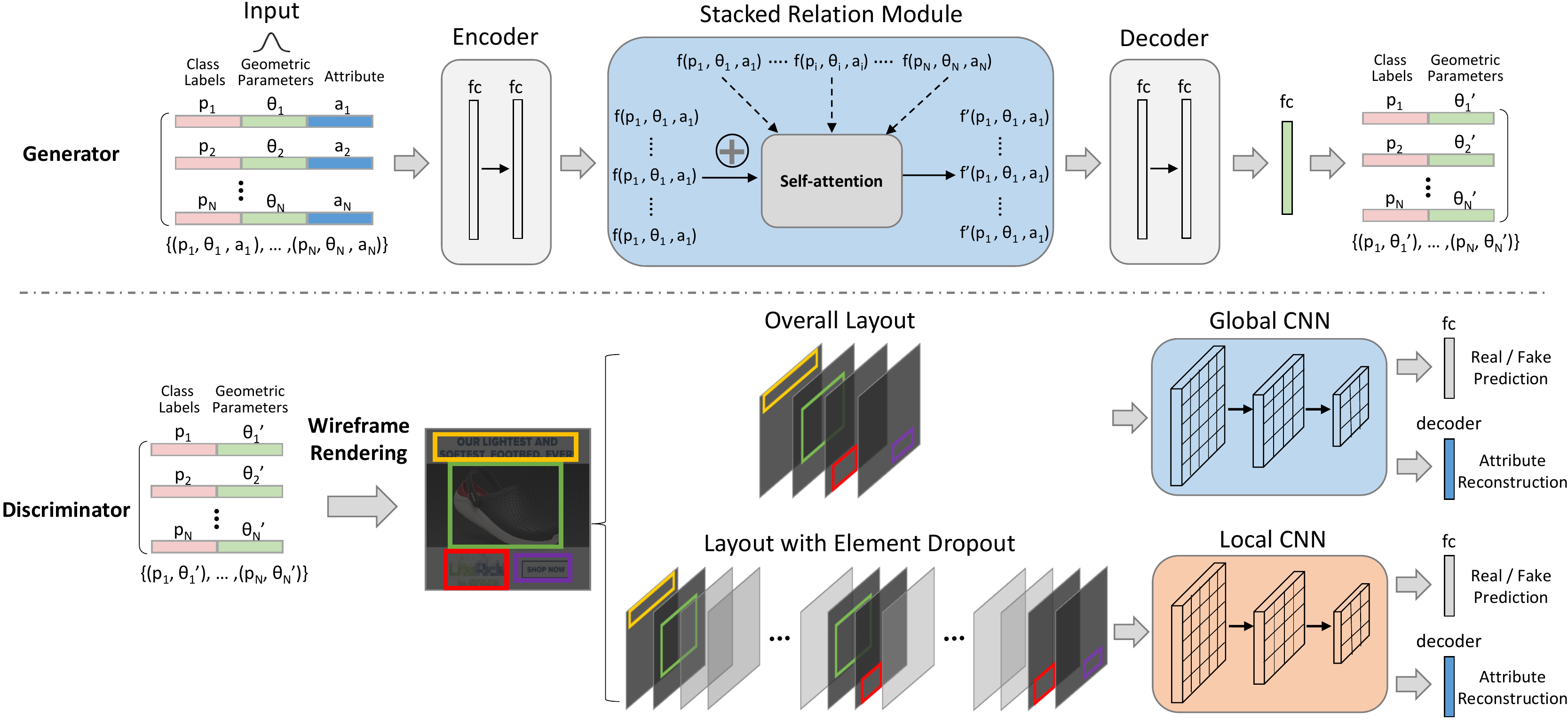}
	\caption{Overall architecture of attribute-conditioned layout GAN. The generator takes graphic elements with class labels, random geometric parameters and attributes as input. An encoder followed by a stacked relation module embeds the input and contextually refines the embedded features for each element. The refined features are then decoded back to geometric parameters to assemble layouts. The discriminator takes generated structured data as input and renders all elements and a random partial list of them into two wireframe images, upon which two CNNs learn to capture both global and local behaviours of elements respectively. The discriminator is further tasked for attribute reconstruction to enhance discrimination.}
	\label{fig:method-framework}
	\vspace{-2ex}
\end{figure*}

\section{Attribute-Conditioned Layout GAN}

\subsection{Design Representation}
Instead of generating pixel-level images as a layout representation, the model synthesizes a graphic layout comprised of a set of N design elements represented as $\left \{(\bm{p_{1}}, \bm{\theta_{1}}),\cdots, (\bm{p_{N}}, \bm{\theta_{N}})\right \}$, where $\bm{p}$ is a vector of the class label and $\bm{\theta} = \left[ \theta^{1}, ..., \theta^{m}\right]$ denotes a set of $m$ geometric parameters that can be in various forms for different graphic layouts~\cite{li2018layoutgan}. We assume an element is a rectangle, or bounding box represented by its normalized center coordinates, width and height, i.e., $\bm{\theta} = \left[x^C, y^C, w, h\right]$, which is very common in various graphic designs. 

\subsection{Attribute Description}
A good professional design, as exampled in Figure~\ref{attr-org}, is built on a layout in which all design elements are arranged in proper sizes and locations according to their content-based attributes such as area, aspect ratio and reading-order.

\subsubsection{Expected Area}
The size of a design element in the layout usually highly depends on its content. For example, for a text element, the area of the generated bounding box should match with the text length, so that the target text can be rendered into the generated region with proper font size to ensure readability and aesthetic soundness. Otherwise, as shown in Figure~\ref{attr-area}, generating a small bounding box (in yellow) for a large amount of text would make the rendered text too small, thus ruining a design. The expected area, determined by elements' content, is thus an essential attribute to be considered for layout design.

\subsubsection{Aspect Ratio}
For an image element, the aspect ratio of the generated bounding box is supposed to be fixed. Given a generated bounding box of an arbitrary aspect ratio, current methods mostly fit the image into the generated region either via cropping or via warping. However, cropping the content could result in information loss and the warped content may result in unwanted geometric distortion, shown as the red bounding box in Figure~\ref{attr-ratio}. The aspect ratio of the image element is thus a necessary attribute to be preserved in the output layout. 

\subsubsection{Reading-order}
\label{sec:reading-order}
For a good design layout, information is presented smoothly based on the reading-order of design elements, as shown in Figure~\ref{attr-org}. In contrast, a layout with disordered elements would not convey information as clearly, as depicted in Figure~\ref{attr-order}. We use simple heuristics by following the left-to-right and top-to-down observation order commonly used in banner ads. In particular, we first compute the distance to the layout origin, i.e., the upper-left corner of the layout canvas, for element $i$ as
$d_{i} = \sqrt{{x_i^L}^2 + {y_i^T}^2}$,
where ($x_i^L, y_i^T$) denotes the upper-left coordinates of the predicted bounding box. Then we sort the computed distance for all elements in ascending order, and take the sorted order of element $i$ as its reading-order $o_i$. Such reading-order is supposed to be retained when adjusting an existing layout to new sizes. It should be taken as an additional attribute when needed. 

\subsection{Attribute Conditioning}
We introduce element attributes in both the generator and the discriminator built upon LayoutGAN~\cite{li2018layoutgan} for layout design.

\subsubsection{Generator Architecture}
\label{sec:gen_arch}
As illustrated in Figure~\ref{fig:method-framework}, the generator takes as input a layout $\bm{z} = \left \{(\bm{p_{1}}, \bm{\theta_{1}}, \bm{a_{1}} ),\cdots, (\bm{p_{N}}, \bm{\theta_{N}}, \bm{a_{N}})\right \}$ consisting of $N$ initial graphic elements, each of which is represented by a pre-defined class label $\bm{p}$ (one hot vector), randomly sampled geometric parameters $\bm{\theta}$, and expected element attributes $\bm{a}$. 
The generator learns to capture and adjust the relations among all elements jointly through the relation module (implemented as self-attention in Wang et al.~\cite{wang2018non}). It does so by embedding the feature of each graphic element as a function of its relations with all the other elements in the design, and outputs $\bm{p}$ and contextually refined $\bm{\theta'}$ to assemble graphic layouts $\bm{G}(\bm{z}) = \left \{(\bm{p_{1}}, \bm{\theta_1'}),\cdots, (\bm{p_N}, \bm{\theta_N'})\right \}$ similar as real ones $\bm{x}$.
Differently from LayoutGAN, we further incorporate an attribute vector $\bm{a}$ as input conditions to generate attribute conditional layouts. In our task, $\bm{a_i}=[s_i,r_i,d_i]$ represents the expected area, aspect ratio and reading-order of element $i$, which determines the sizes and relative locations of the generated bounding box.
Here, $r_i$ equals the expected aspect ratio (bounding box height divided by width) if element $i$ is supposed to be ratio-fixed, and $0$ otherwise. Denote $h_i'$ and $w_i'$ as the height and width of the bounding box originally predicted by the generator, respectively. To strictly obey the aspect ratio constraint introduced by $r_i$, the final output bounding box height $\hat{h}_i'$ can be computed with:
\begin{equation}
\hat{h}_i' = \bm{1}[r_i=0]h_i' + r_iw_i',
\end{equation} 
where the Iverson bracket indicator function $\bm{1}[r_i=0]$ equals $1$ if $r_i=0$, and $0$ otherwise. 

\subsubsection{Discriminator Architecture}
The discriminator renders synthesized structured data into wireframe images using a differentiable wireframe rendering layer~\cite{li2018layoutgan}, and learns to differentiate between synthesized and real layouts by capturing the relations among elements using CNNs from the visual domain. To discriminate attribute conditional layouts, we further task the discriminator for attribute reconstruction by introducing an additional decoder (implemented as fully connected layers) as inspired by Odena et al.~\cite{odena2017conditional}. As element-wise information has been lost during wireframe rendering, we force the discriminator to reconstruct class-wise attribute information instead, i.e., the total area for each element type in the input layout. Specifically, the total area for element type $c$ can be computed with:
$S_c = \sum_{i=1}^{N}p_{i,c}s_i$,
where $p_{i,c}$ is the class probability on element type $c$ for element $i$. 
As a result, the discriminator outputs both a probability of the input belonging to real layouts, and an estimated area distribution over all element types.

\subsection{Element-Dropout Wireframe Rendering}
We adopt CNNs to better capture visual properties of an layout by rasterizing graphic elements to 2D wireframe images. To demonstrate this, we consider rasterizing a design comprised of $N$ elements parameterized as $\left \{(\bm{p_1}, \bm{\theta_1}), ..., (\bm{p_N}, \bm{\theta_N})\right \}$ to a multi-channel output image $\bm I$ of dimension $W \times H \times M$, where each channel corresponds to one of $M$ element types, and $W$ and $H$ denote the width and height of the layout canvas in pixels respectively. Each design element can be rendered into its own grayscale image $\bm F_{\bm{\theta}} (x,y)$ of dimension $W \times H$, as detailed in Li et al.~\cite{li2018layoutgan}. The class activation vector for pixel $(x,y)$ on $\bm{I}$ can be computed with:
\begin{equation}
\bm I(x,y,c) = \max_{i\in[1...N]} p_{i,c} {\bm F_{\theta_i}(x,y)}.
\end{equation}

The wireframe discriminator mainly perceives global patterns of all elements. However, local behaviours such as alignment and non-overlapping are also important for a good design. To better capture those local rules, we further introduce element dropout strategy by randomly removing some input elements during the rendering process. Denote $\bm r$ as a N-dimension vector of independent Bernoulli random variables, each of which has a probability $b$ of being $1$. This vector is sampled and multiplied with the rendered grayscale image $\bm{F_{\theta}}(x,y)$ to produce another rendered image $I'$: 
\begin{equation}
r_{i} \sim Bernoulli(b),
\label{eqn:bernoulli}
\end{equation}    
\begin{equation}    
\bm {I'}(x,y,c) = \max_{i\in[1...N]} r_{i}p_{i,c} {\bm F_{\theta_i}(x,y)}.
\end{equation}

Figure~\ref{fig:method-framework} depicts the architecture of the discriminator, which consists of both a global and a local branch. The global branch renders all input elements to wireframe images and uses CNNs to measure and optimize global patterns. Meanwhile, the local branch randomly removes some elements during rendering and adopts additional CNNs to focus on local patterns by looking at partial lists of elements. The discriminator is thus able to well capture both the global and the local layout patterns. 

\subsection{Discriminator Optimization}
Denote $\bm{D_{\Theta_a}}$ and $\bm{D_{\Theta_a'}}$ as the global and the local branch with parameter $\bm{\Theta_a}$ and $\bm{\Theta_a'}$ respectively, and $\bm{G_{\Theta_g}}$ as the generator with parameter $\bm{\Theta_g}$. The objective function of the discriminator consists of two terms. One is the log-likelihood for the input of being real layouts:
\begin{equation}
\begin{split}
\bm{L_a} = &-\log \bm{D_{\Theta_a}}(\bm{x}) -\log (1-\bm{D_{\Theta_a}}(\bm{G_{\Theta_g}}(\bm{z})))\\
& -\log \bm{D_{\Theta_a'}}(\bm{x}) -\log (1-\bm{D_{\Theta_a'}}(\bm{G_{\Theta_g}}(\bm{z}))).
\end{split}
\end{equation} 
The other one is the $L_1$ distance between the predicted and the real area distribution over all element types:
\begin{equation}
\bm{L_r} = \sum_{c=1}^{M}\left | S_c - S_c' \right |,
\end{equation} 
where $S_c$ and $S_c'$ represent the real and the predicted area for element type $c$ respectively. The discriminator is trained to minimize $\bm{L_a} + w_r\bm{L_r}$, where $w_r$ is a preset loss weight. 

\subsection{Generator Optimization}
\label{sec:gen_optimize}
We introduce several losses based on different design principles for layout optimization. 

\subsubsection{Adversarial Loss}
The quality of the layout depends, among other things, on overall composition on the page (e.g., use of positive and negative space, balance, etc). Manually designing losses to optimize such patterns is not trivial, and may lead a very complex energy function to optimize. We thus adopt adversarial training to learn target design patterns from real data automatically.
Taking the generated layout as input, the discriminator predicts its probability of belonging to a real layout by assessing both the global and the local behaviours of design elements. 
Trying to fool the discriminator produces an adversarial loss $L_{adv}$, which encourages the generator to synthesize layouts similar as real ones:
\begin{equation}\label{eqn:loss}
{L_{adv}} = -\log \bm{D_{\Theta_a}}(\bm{G_{\Theta_g}}(z))-\log \bm{D_{\Theta_a'}}(\bm{G_{\Theta_g}}(z)).
\end{equation}

\subsubsection{Margin Area Loss}
\label{sec:area_loss}
We control the area of the generated bounding box for different elements by feeding their expected area attributes to the generator. 
We denote $s_i$ and $s_i'$ as the input expected area and the area of the predicted bounding box for element $i$ respectively.
$s_i$ and $s_i'$ are expected to be close in value, while remaining some degree of flexibility to better balance different design principles for layout optimization.
We thus introduce a margin area loss:
\begin{equation} 
L_{area} = \sum_{i=1}^{N}\bm{k}\left ( \frac{\left |s_i'-s_i\right |}{s_i} - \alpha \right ),
\end{equation} 
where $\bm{k}(x) = \max (0,x)$ (implemented as ReLU activation), and $\alpha \geq 0$ is a preset parameter to control the allowable relative area difference, which is set as $0.3$ in our experiments.

\subsubsection{Overlapping Loss}
Well-designed layouts typically avoid overlapping elements.
We offer an overlapping loss to penalize overlapping between any element pairs:
\begin{equation}
L_{over} = \sum_{i=1}^{N}\sum_{\forall j \neq i}\frac{s_i\cap s_j}{s_{i}},
\label{eqn:loss-iou}
\end{equation} 
where $s_i\cap s_j$ denotes the overlapping area between element $i$ and $j$.

\begin{figure*}
	\centering
	\includegraphics[width=2.0\columnwidth]{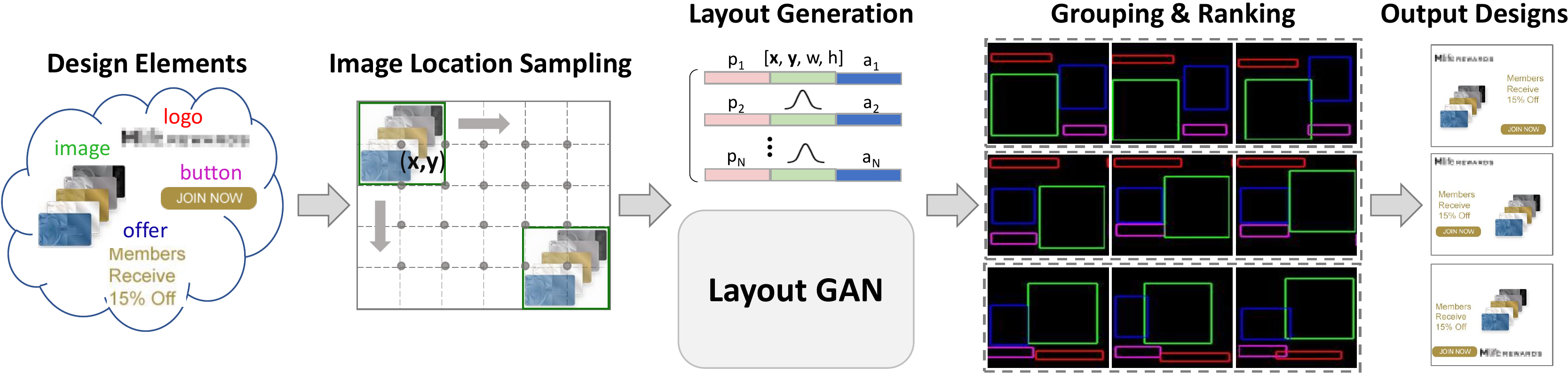}
	\caption{Overall pipeline for automatic advertisement design. Given a set of design elements, the system first samples a number of locations for the product image, and then feeds the sampled locations and elements' attributes to layout GAN to generate a number of layouts, which are further grouped and ranked for final advertisement rendering and recommendation.}
	\label{fig:design-generation}
	\vspace{-2ex}
\end{figure*}

\subsubsection{Alignment Loss}
The alignment of design elements is crucial to how viewers perceive a layout. Adjacent elements are usually in six possible alignment types: Left, X-center, Right, Top, Y-center and Bottom aligned. Denote $\bm{\theta} = (x^{L},y^{T},x^{C},y^{C},x^{R}, y^{B})$ as the top-left, center and bottom-right coordinates of the predicted bounding box, we encourage pairwise alignment among elements by introducing an alignment loss: 
\begin{equation}
L_{alg} = 
\sum_{i=1}^{N}\min \begin{pmatrix}
\bm{g}(\Delta x_{i}^{L}),\ \bm{g}(\Delta x_{i}^{C}),\ \bm{g}(\Delta x_{i}^{R}), \\ 
\bm{g}(\Delta y_{i}^{T}),\ \bm{g}(\Delta y_{i}^{C}),\ \bm{g}(\Delta y_{i}^{B})
\end{pmatrix},
\label{eqn:loss-align}
\end{equation} 
where $\bm{g}(x) = -log(1-x)$, and $\Delta x_{i}^{*} (*=L,C,R)$ is computed as:
\begin{equation}
\Delta x_{i}^{*} = \min_{\forall j \neq i} \left | x_{i}^{*} - x_{j}^{*} \right |.
\end{equation}
$\Delta y_{i}^{*} (*=T,C,B)$ can be computed similarly.

\subsubsection{Order Loss}
\label{sec:order-loss}
The reading-order of elements is expected to be retained when adjusting an existing layout to new sizes. 
Following the left-to-right and top-to-down observation order commonly used in banner ads, a design element closer to the layout origin, i.e., the upper-left corner of the layout canvas, is expected to have a prior reading-order. As defined in Section~\ref{sec:reading-order}, denote $o_i$ as the pre-defined reading-order of element $i$ and $d_i$ as its distance to the layout origin. If element $i$ has a prior reading-order compared to element $j$, i.e., $o_i < o_j$, the former is supposed to be closer to the layout origin than the latter accordingly, i.e., $d_i < d_j$. We thus introduce an order loss to penalize any element pairs whose relative distances to the layout origin are against their pre-defined reading-orders during layout adjustment:
\begin{equation}
L_{ord} = \sum_{i=1}^{N}\sum_{j=1}^{N}\bm{1}[o_{i}<o_{j}]\bm{k}(d_{i}-d{j}),
\end{equation} 
where $\bm{k}(x) = \max (0,x)$.
$\bm{1}[o_{i}<o_{j}]$ equals $1$ when $o_{i}<o_{j}$, and $0$ otherwise.

The final objective function is a weighted sum of all the losses above, whose weights are set as $0.6$, $4.0$, $8.0$, $20.0$ and $20.0$ respectively in our experiments. We found the model is not very sensitive to different settings of these weights.

\section{Applications}
A possible application of our model is in designing advertisements.
Given a number of design elements of different types such as the logo, product image, headline, button, offer and disclaimer, the task of advertisement design is to specify the locations and sizes of different elements to form a well-organized layout.
Figure~\ref{fig:design-generation} illustrates our overall pipeline for automatic advertisement design, which consists of three basic steps: image location sampling, attribute-conditioned layout generation and layout grouping and ranking. We further illustrate how to adjust existing layouts to new sizes by incorporating additional reading-order constraints. 

\emph{Image Location Sampling}
Advertisement designs always use the product image as a starting point.
Its location and size could greatly determine the layout style, and are thus supposed to be specified as design prior so that other elements can be arranged appropriately.
To this end, the geometric parameters of the product image can be given as input conditions, while those for other element types are randomly sampled from a Gaussian distribution and to be refined accordingly to assemble a reasonable layout.
Specifically, the width and height of the output bounding box for a given product image is computed from its expected area and aspect ratio.
As for center coordinates, we apply grid sampling to obtain a number of candidate locations uniformly distributed inside the layout canvas, at which the product image can be accommodated. 

\begin{figure}[t]
	\centering
	\includegraphics[width=1.0\columnwidth]{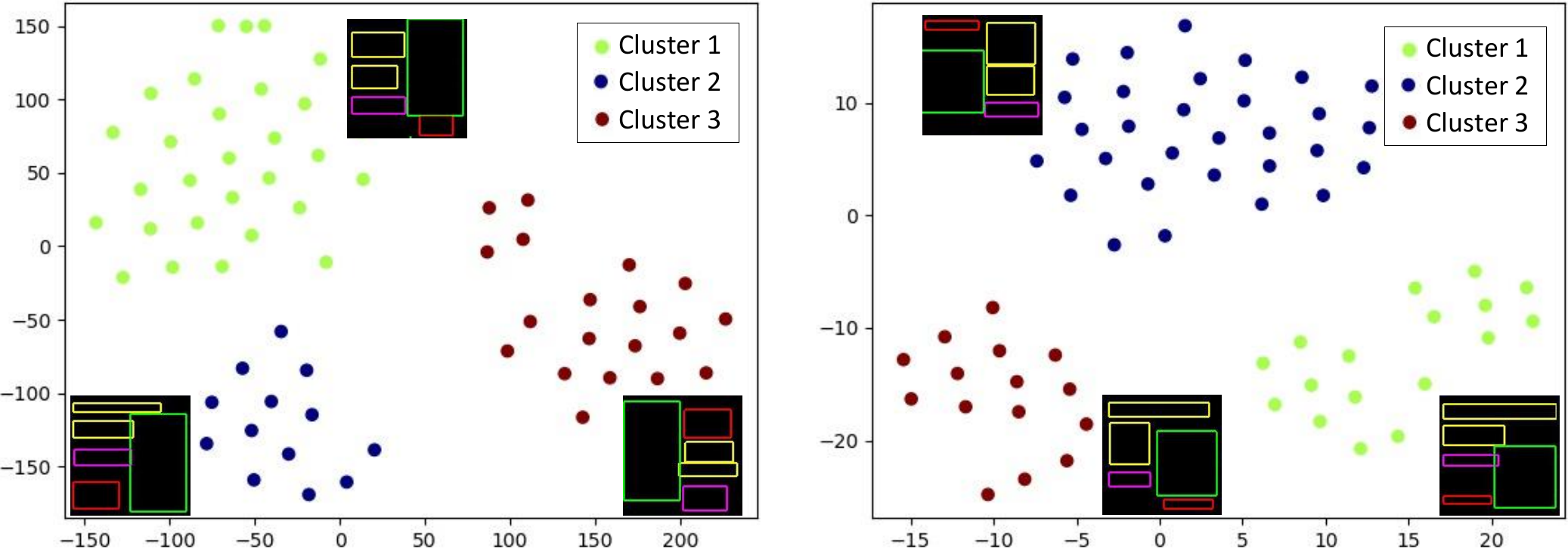}
	\caption{Visualizing layout features using t-SNE. Given a set of design elements, the model generates a number of layouts with different patterns, which can be grouped into clusters based on the feature similarity. The representative layout for each clustered group is displayed nearby.}
	\label{fig:visual-cluster}
	\vspace{-2ex}
\end{figure}

\begin{figure*}[t]
	\centering
	\includegraphics[width=1.8\columnwidth]{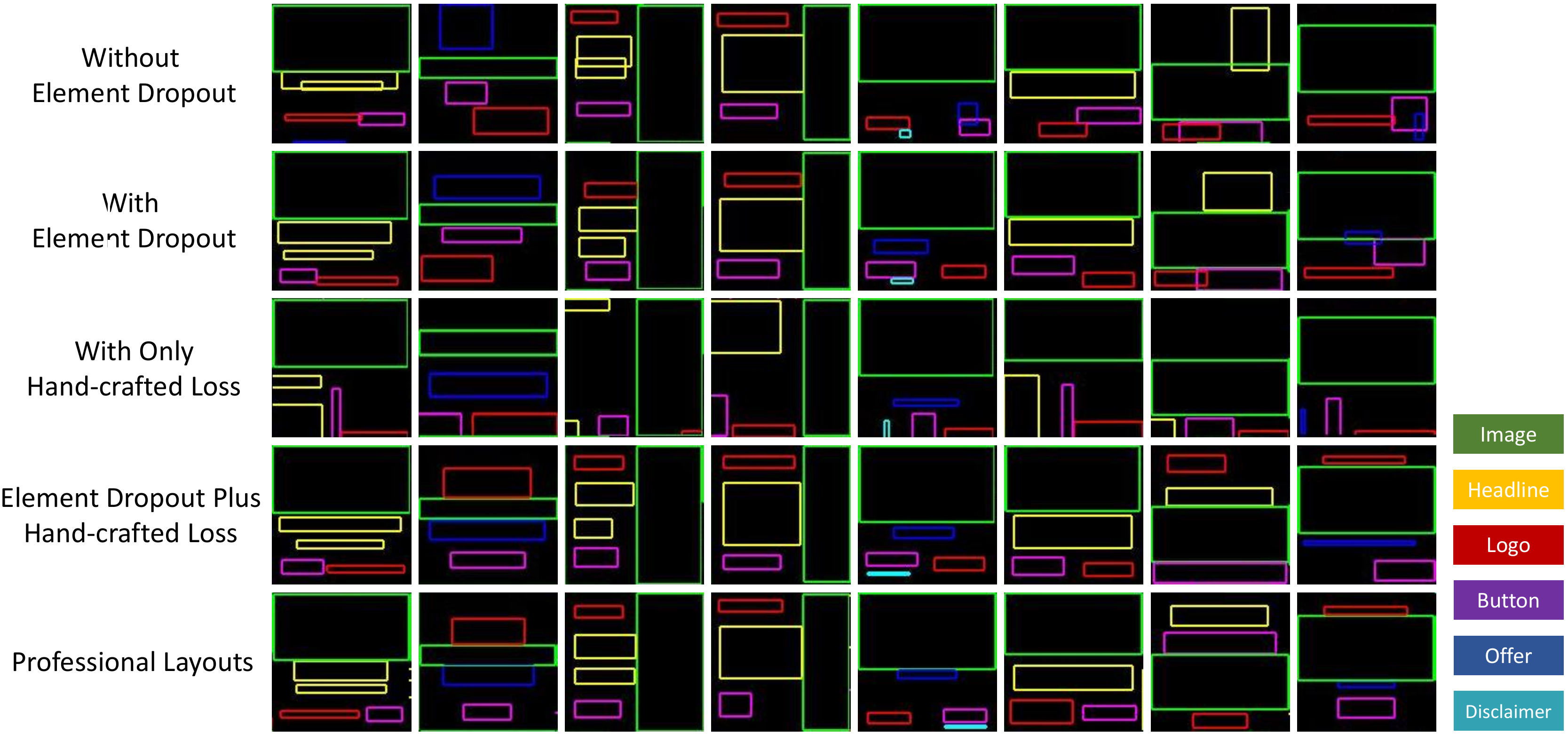}
	\caption{Qualitative comparisons of generated layouts from different model variants.}
	\label{fig:visual-dropout}
\end{figure*}

\emph{Attribute-conditioned Layout Generation}
The layout arrangement of different design elements would highly depend on their content-based attributes. For text elements such as headlines, buttons, offers and disclaimers, the expected area determined by the text length is an essential condition for generating bounding boxes of proper scale so that text can be rendered with appropriate font size. For image elements such as logos, the aspect ratio is supposed to be preserved, and thus should be taken as an additional input attribute. Given the above content-based attributes, the model produces a number of layouts based on different sampled locations for the product image. We trained three models for generating layouts of different aspect ratios, i.e., portrait, square and landscape.

\emph{Grouping and Ranking}
A concrete automatic design system should be able to assess the quality of different generated layouts for recommendation. Given a number of generated layouts, we first feed them into the trained discriminator and extract features from the last convolutional layer. We visualize the features for different layouts using t-SNE~\cite{maaten2008visualizing}. As shown in Figure~\ref{fig:visual-cluster}, presence of clusters can be clearly found, which provides evidence that there exist similar patterns among different generated layouts. We thus divide all generated layouts into groups based on the feature similarity using k-means clustering, and assess the quality of each generated layout using the following cost function:
\begin{equation}
E = \omega_{adv} L_{adv} + \omega_{over} L_{over} + \omega_{alg} L_{alg},
\end{equation}
where $\omega_{adv}$, $\omega_{over}$ and $\omega_{alg}$ denote the weights for the adversarial, overlapping and alignment loss terms mentioned in Section~\ref{sec:gen_optimize} respectively. These weights are set as the same values as used for model optimization. Finally, we rank layouts in each clustered group according to their computed cost values in descending order, and output the first-ranked layout in each group as the recommended result. 

\emph{Graphic Layout Adjustment}
As designs are usually presented in various displays, we sometimes need to adjust existing generated layouts to new sizes and aspect ratios while retaining elements' reading-orders. Though size free, the layout generation pipeline can only produce layouts of certain aspect ratios. We thus further train another generative model for layout adjustment by including reading-order as additional input attributes and by applying the order loss in Section~\ref{sec:order-loss} to retain such information in the adjusted layouts. In addition, we set the location of the product image to be predictable without input prior to better meet the reading-order constraint. For inference, given a generated layout to be adjusted, we first extract elements' attributes and apply transformation to the aspect ratio of the logo and the product image according to the source and the target layout canvas size. Then, we feed the transformed attributes to the trained layout adjustment model to generate intermediate layouts, which are further resized to the target canvas size as final adjustment results.

\begin{table}\renewcommand\arraystretch{1.1}
	\centering\tabcolsep=0.3cm
	\caption{Spatial analysis of layouts from different models.}
	\begin{tabular}{ccc}
		\toprule
		Methods                                & Overlap     & Alignment \\  
		\midrule		
		Without element dropout                   & 0.68  & 0.09 \\ 
		With element dropout                      & 0.39  & 0.08 \\ 
		Element dropout + overlapping loss    & 0.09  & 0.08 \\ 
		Element dropout + alignment loss      & 0.36  & 0.04 \\ 
		Element dropout + both losses         & 0.08  & 0.04 \\ 	 
		Professional layouts                      & 0.00  & 0.04 \\ 			
		\bottomrule
	\end{tabular}
	\label{tab:ele-dropout}
	\vspace{-2ex}
\end{table}

\section{Experiments}
The implementation is based on TensorFlow~\cite{abadi2016tensorflow}. The network parameters are initialized from zero-mean Gaussian with standard deviation of $0.02$. All network parameters are optimized using Adam~\cite{kingma2014adam} with a fixed learning rate of $0.00001$. In our experiments, we focus on advertisement layouts with two to six
bounding boxes that may belong to six most frequent element types appeared in real designs, as mentioned earlier. For design data, we collect totally around $17,000$ advertisement layouts designed by professional designers, which are of three different aspect ratios, i.e., portrait, square and landscape. We randomly split these collected layouts into a training and a testing set and extract elements’ attributes in each layout as design conditions.

\begin{table}\renewcommand\arraystretch{1.1}
	\centering\tabcolsep=0.75cm
	\caption{Comparisons of using different dropout probabilities.}
	\begin{tabular}{ccc}
		\toprule
		Probability         & Overlap     & Alignment \\ 
		\midrule		
		$b=0.25$             & 0.13  & 0.05 \\ 
		$b=0.50$             & 0.08  & 0.04 \\ 
		$b=0.75$             & 0.09  & 0.04 \\ 
		$b=1.00$             & 0.11  & 0.05 \\ 
		\bottomrule
	\end{tabular}
	\label{tab:ele-dropout-prob}
\end{table}

\begin{table}\renewcommand\arraystretch{1.1}
	\centering\tabcolsep=0.7cm
	\caption{Comparisons of using different attribute conditioning methods.}
	\begin{tabular}{ccc}
		\toprule
		Conditioning         & Overlap     & Alignment \\ 
		\midrule
		$w_r=0.0$             & 0.10  & 0.05 \\ 
		$w_r=0.5$             & 0.08  & 0.04 \\ 	
		$w_r=1.0$             & 0.12  & 0.04 \\ 
		$Feeding$             & 0.15  & 0.05 \\ 
		\bottomrule
	\end{tabular}
	\label{tab:attribute-rect}
	\vspace{-2ex}
\end{table}

\subsection{Ablation Studies}
We investigate the effectiveness of different model components.

\subsubsection{Element Dropout Strategy}
We incorporate a local branch with element dropout into the discriminator to better capture local patterns of layouts. For validation, we compare the proposed model with the model variant without such local branch. Note that in this experiment, we do not apply the overlapping and the alignment loss when optimizing the above two models to purely compare discriminator behaviors. The first two rows in Figure~\ref{fig:visual-dropout} provide some example layouts generated by the above two models respectively. We also retrieve similar professional layouts that best match the input elements' attributes for reference, as shown in the last row. One can see that the additional local branch with element dropout helps better optimize elements' pairwise relations, thus alleviating overlapping and misalignment problems. 

For quantitative evaluation of the generated layouts, we propose two metrics, overlapping and alignment index (the lower the better), as formulated in Equation~\ref{eqn:loss-iou} and~\ref{eqn:loss-align} respectively. The first two rows in Table~\ref{tab:ele-dropout} provide quantitative comparisons of both models. We also include metric values for professional layouts in the last row for reference. They demonstrate that incorporating element dropout achieves lower values in both metrics, validating its superiority in optimizing local layout patterns. In addition, we compare results of adopting different dropout probabilities, referred to as $b$ in Equation~\ref{eqn:bernoulli}. Table~\ref{tab:ele-dropout-prob} shows both metrics achieve the lowest when $b$ equals $0.5$. We use such settings in all experiments. 

\begin{figure}
	\centering    
	\begin{subfigure}{0.24\textwidth}
		\centering
		\includegraphics[width=1.03\columnwidth]{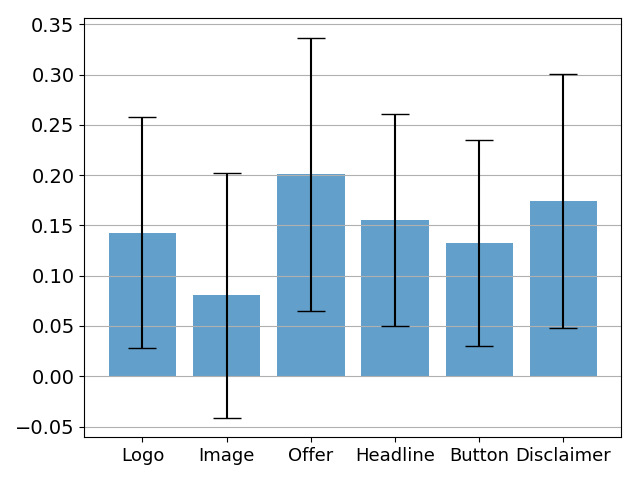}
		\caption{Relative area difference.}
		\label{fig:statistics_areadiff}
	\end{subfigure}	
	\begin{subfigure}{0.235\textwidth}
		\centering
		\includegraphics[width=1.06\columnwidth]{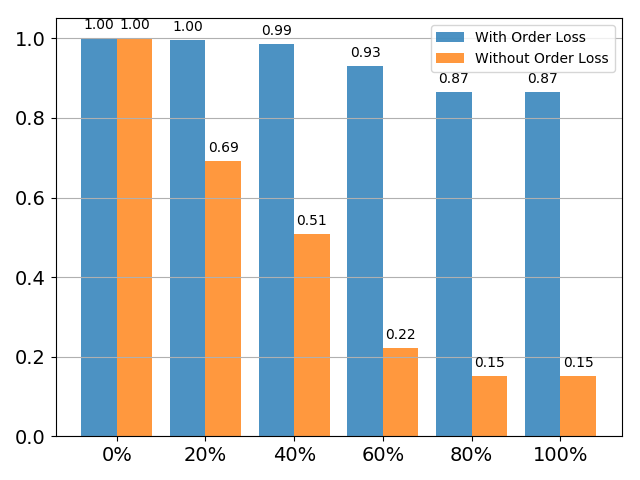}
		\caption{Reading-order retention.}
		\label{fig:order_per}
	\end{subfigure}	
	\caption{Analysing the capability of preserving attributes.}
	\vspace{-2ex}
\end{figure}

\subsubsection{Hand-crafted Loss}
Considering the complexity of layout patterns, we further introduce hand-crafted losses, i.e., the overlapping and the alignment loss for optimization. For validation, we compare the results from model variants trained with only the adversarial loss, with only the hand-crafted loss and with both loss terms. As depicted in the second to fourth rows of Figure~\ref{fig:visual-dropout}, optimized layouts by a single adversarial loss suffer some degree of misalignment and overlapping. While using only the hand-crafted loss leads to degeneration. By combing both loss terms together, the model produces satisfactory results. 

Similar conclusions can be drawn from quantitative comparisons. As shown in the middle four rows of Table~\ref{tab:ele-dropout}, introducing hand-crafted loss terms incrementally based on adversarial training offers a consistent decrease in both metrics. To prove that introducing the hand-crafted loss would not affect overall layout balance, we further provide a supplementary metric to measure design symmetry in this experiment. Specifically, symmetric balance is a common design principle for organizing elements in conventional layouts. As the real layouts in our collected data mostly contain horizontal symmetry, we adopt the balance formulation inspired by O'Donovan et al.~\cite{o2014learning} to obtain symmetric scores, which compute the fraction of pixels having symmetric counterparts along the horizontal axis in the rendered layout. Incorporating the hand-crafted loss improves the symmetric score from $74.33\%$ to $76.47\%$, which is closer to the score for collected real layouts, i.e., $76.51\%$. This suggests that the proposed hand-crafted loss would not affect global balance of layouts.

\begin{figure}[t]
	\centering
	\includegraphics[width=1.0\columnwidth]{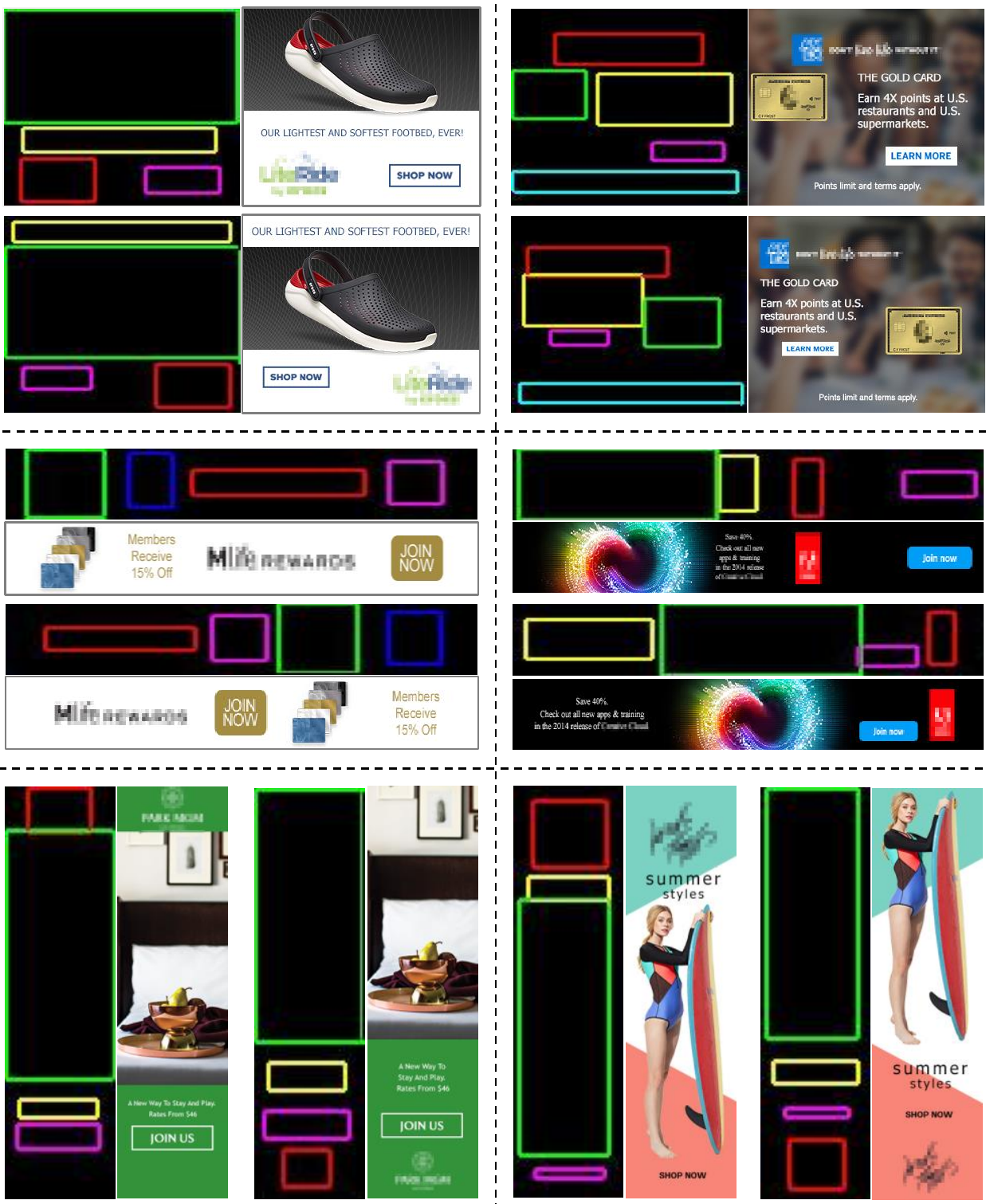}
	\caption{Synthesized layout samples and corresponding rendered designs of different aspect ratios.}
	\label{fig:visual-genlayout}
\end{figure}

\begin{figure}[t]
	\centering
	\includegraphics[width=1.0\columnwidth]{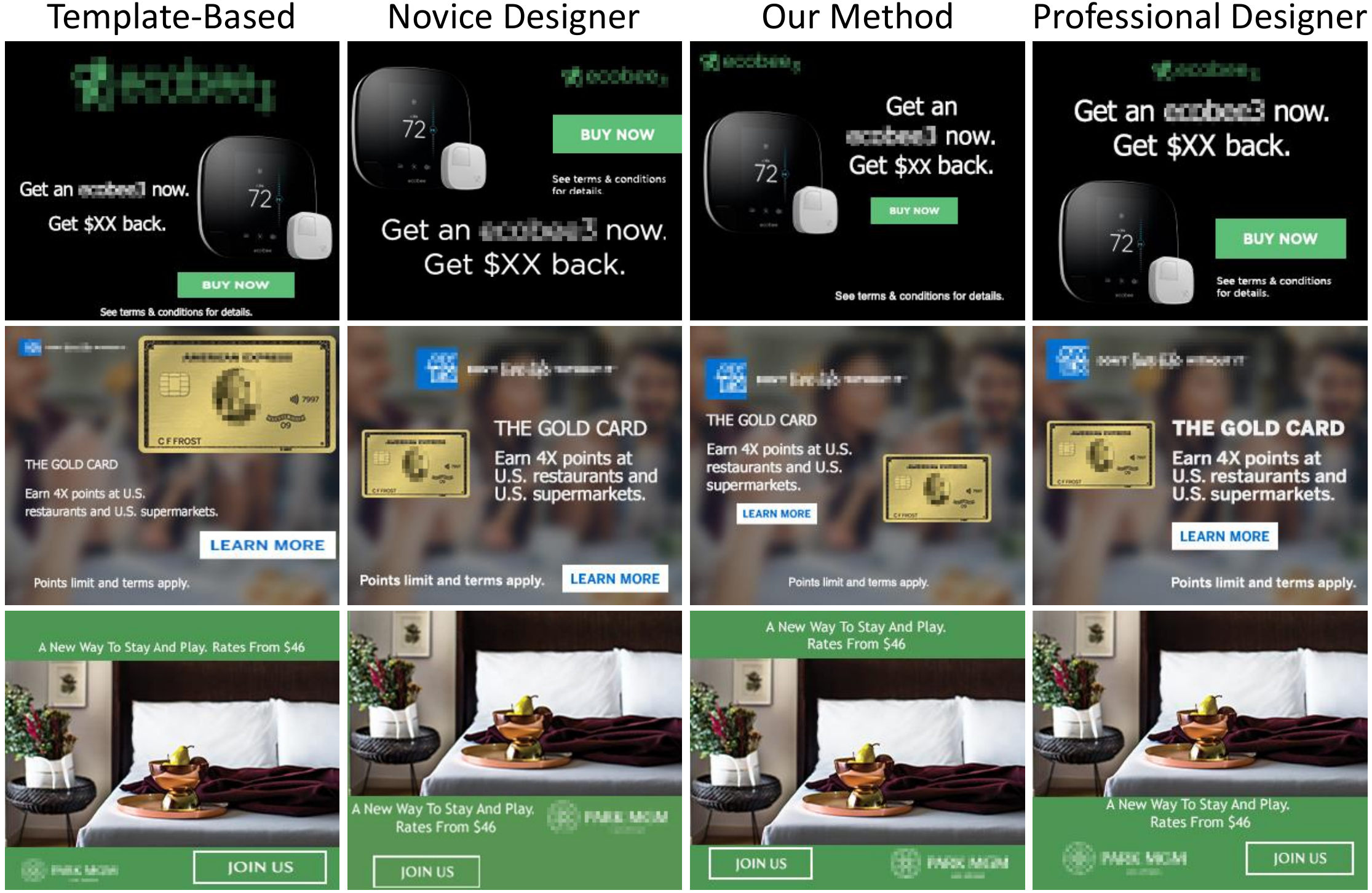}
	\caption{Comparisons of designs from different methods.}
	\label{fig:visual-comparsion}
	\vspace{-2ex}
\end{figure}

\subsubsection{Attribute Reconstruction}
We build the conditional discriminator by forcing it to reconstruct attributes of input layouts. Table~\ref{tab:attribute-rect} provides quantitative comparisons of using different loss weights for the attribute reconstruction task. All models with conditional discriminator ($w_r>0$) achieve lower values in both metrics compared to the model variant without performing attribute reconstruction ($w_r=0$). This suggests that the conditional discriminator effectively leverages attribute information through reconstruction, and thus improves discrimination for attribute conditional layouts. In addition, a proper loss weight, i.e., $w_r=0.5$, contributes to performance boost. Moreover, the last row in Table~\ref{tab:attribute-rect} depicts that incorporating attributes through reconstruction outperforms the alternative manner, i.e., simply feeding attribute information as discriminator input~\cite{radford2015unsupervised}, which further validates our design choice. 

\begin{figure}[t]
	\centering
	\includegraphics[width=0.65\columnwidth]{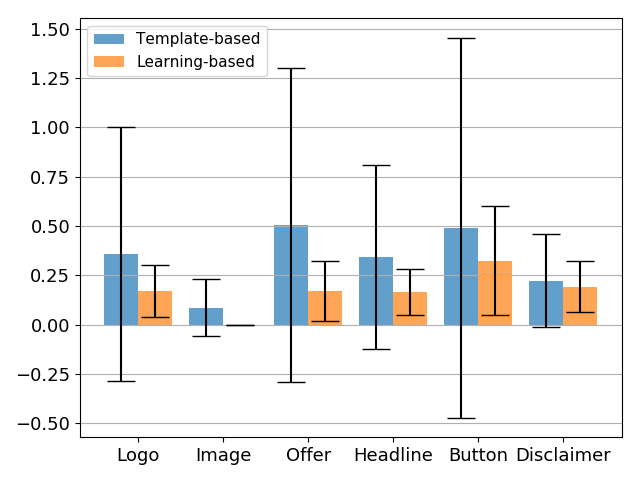}
	\caption{Comparisons of relative area difference.}
	\label{fig:area_comparison}
\end{figure}

\subsubsection{Attribute Preserving}
\label{sec:attr-preserve}
To investigate the effectiveness of preserving input attributes, we analyze the area and the reading-order of output elements to compare them with input attributes based on the layout adjustment task. Specifically, we compute the relative difference between the generated bounding box area and the input expected area for each element, and analyze the mean along with the standard deviation of such area difference for each element type. As shown in Figure~\ref{fig:statistics_areadiff}, the mean area difference for all element types falls below the preset allowable threshold, i.e., $\alpha=0.3$ as described in Section~\ref{sec:area_loss}. To validate retention of the reading-order, we analyze the proportion of generated layouts whose percentage of design elements meeting the input reading-order constraint exceeds specific thresholds. Figure~\ref{fig:order_per} depicts that incorporating the reading-order loss effectively preserves the input reading-order and achieves higher proportion values consistently under different thresholds compared to the model variant without using such loss. The analysis of both area and reading-order retention demonstrates that our model can effectively incorporate and preserve input attributes to generate attribute conditional layouts.

\begin{figure}[t]
	\centering
	\includegraphics[width=0.98\columnwidth]{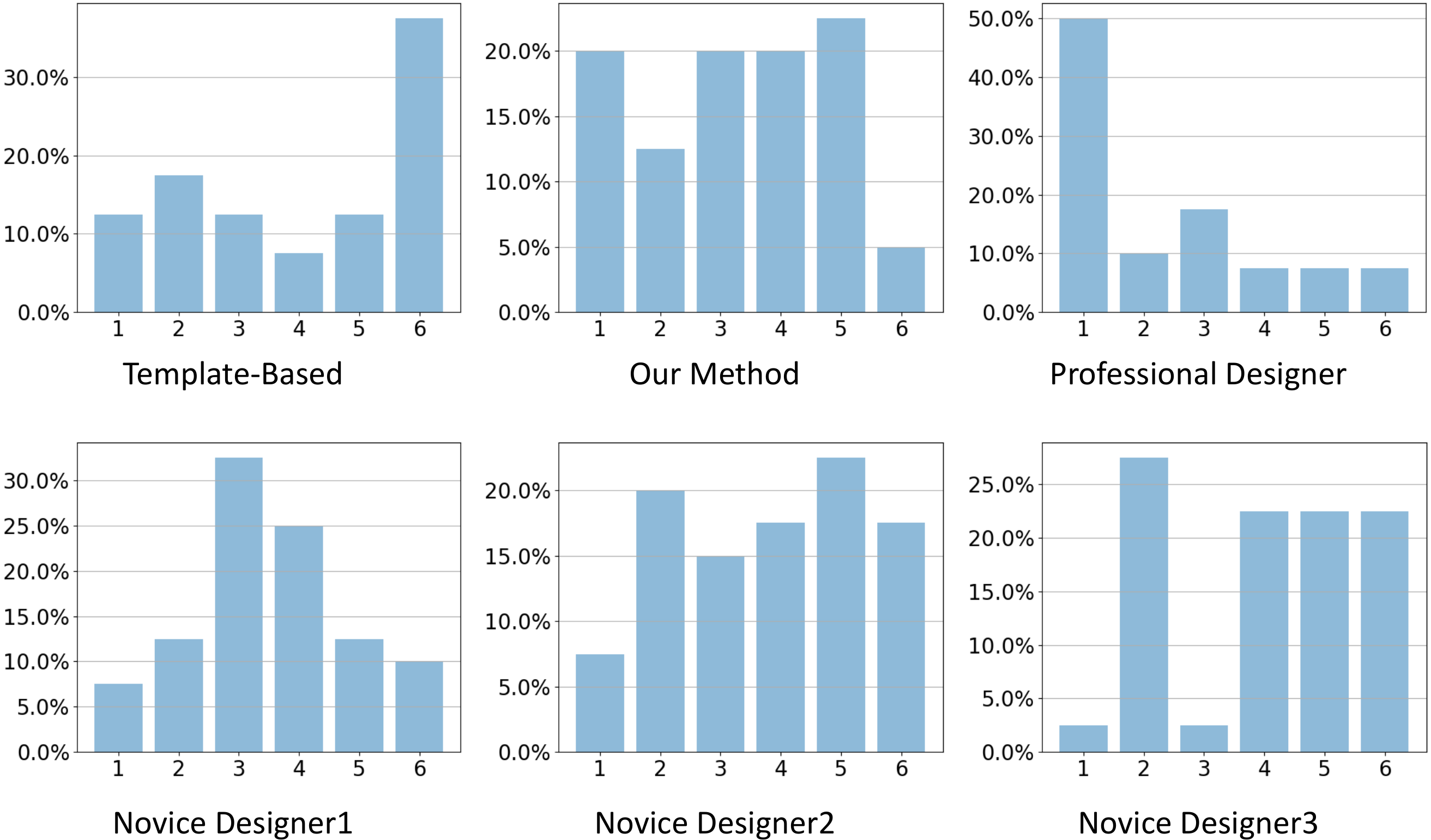}
	\caption{Histograms of rankings for different methods.}
	\label{fig:visual-rank}
	\vspace{-2ex}
\end{figure}

\subsection{Design Results}
For qualitative results, we provide example generated layouts and corresponding rendered advertisements of three aspect ratios (portrait, square and landscape). As illustrated in Figure~\ref{fig:visual-genlayout}, our method can well capture various layout patterns and generate aesthetically pleasing designs of different aspect ratios. In addition, we introduce a template-based baseline for attribute-conditioned layout retrieval. Specifically, given a set of design elements, we first construct a query vector similar as the generator input but comprising only elements' attributes. Then we compute the Cosine similarity between the query vector and the vectors derived from existing professional layouts in the training data, and retrieve the most similar professional layout directly as output. The first and the third column in Figure~\ref{fig:visual-comparsion} provide results from the template-based baseline and our method respectively. The retrieved layouts, though derived from professional designs, poorly meet elements' attribute conditions. As shown in the first two rows, mismatched aspect ratio for logos would ruin a design. 
Figure~\ref{fig:area_comparison} also provides the area difference analysis as introduced in Section~\ref{sec:attr-preserve}.
Our learning-based model can better meet the area constraint compared to the template-based baseline. We further compare our generated results with designs created by professional designers and those by novice designers who lack much design experience and complete designs by simply translating and scaling given elements. Figure~\ref{fig:visual-comparsion} demonstrates that our results are generally better than or at least as good as those produced by novice designers, and sometimes comparable to professional designs.

\begin{table}[t]\renewcommand\arraystretch{1.1}
	\centering\tabcolsep=0.21cm
	\caption{Average rankings for different methods.}
	\begin{tabular}{ccc}
		\toprule
		Methods                  & Ranks (Normal) & Ranks (Professional) \\ 
		\midrule
		Template-based           & 4.10 & 4.03 \\	
		Novice Designer          & 3.58 & 3.78 \\			
		Our Method               & 3.48 & 3.28 \\
		Professional Designer    & 2.66 & 2.35 \\			
		\bottomrule
	\end{tabular}
	\label{tab:usr-gen}
\end{table}

\begin{figure}[t]
	\centering
	\includegraphics[width=1.0\columnwidth]{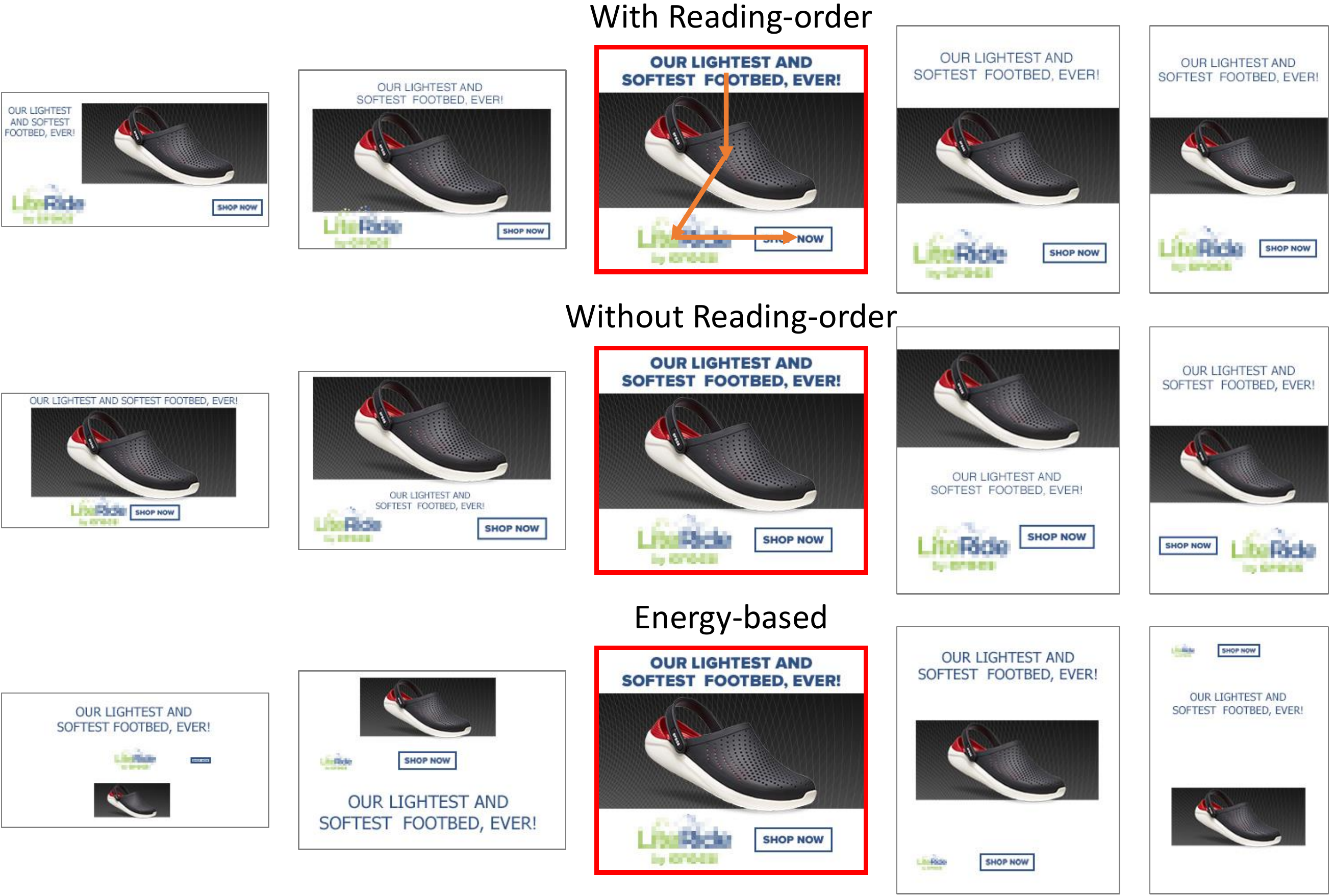}
	\caption{Comparisons of retargets from different methods.}
	\label{fig:visual-retargets}
	\vspace{-1ex}
\end{figure}

For quantitative evaluation, we conduct a user study to subjectively evaluate results from different methods mentioned above. The user study involves two user groups, i.e., $22$ normal users with certain design background and $2$ professional users with rich design experience (different from designers creating the professional designs to be evaluated). Specifically, we collect $20$ sets of design elements and compare six design results for each set, which are produced by our method, the template-based baseline, professional designers and three novice designers, respectively. A subject is asked to rank the given six designs in descending order by following two criteria: 1) whether they are aesthetically displeasing, 2) whether they can clearly convey information. Figure~\ref{fig:visual-rank} depicts the histogram of rankings for designs from different methods and Table~\ref{tab:usr-gen} provides computed average rankings (the lower the better). One can see that the ranking results from the normal and the professional users are consistent. Concretely, though not at the level of professional designs, 
our automatic designs achieve better rankings compared to designs produced by the template-based baseline and also those by novice designers.
This suggests that our generated designs are generally preferred by participants.  

We further adapt our method to adjusting existing layouts to new sizes while retaining elements' reading-orders. 
The first two rows in Figure~\ref{fig:visual-retargets} provide retargets produced by our method and the model variant without considering the reading-order constraint respectively when given a specific design (middle in red) as input.
We also compare with results from the energy-based model proposed by O'Donovan et al.~\cite{o2014learning}, as shown in the third row. 
One can see that our method can properly specify the locations and sizes of different elements while retaining original reading-orders, thus generating more appealing and adequate retargets compared to the other two counterparts.

At last, we compare the average inference time of producing an automatic graphic layout by using different methods, i.e., the template-based baseline, the energy-based model~\cite{o2014learning} and our method. As shown in Table~\ref{tab:test-time}, our method is greatly faster or at least comparable in terms of speed compared to the alternatives, while creating better designs.

\begin{table}[t]\renewcommand\arraystretch{1.1}
	\centering\tabcolsep=0.12cm
	\caption{Comparisons of average inference time.}
	\begin{tabular}{cccc}
		\toprule
		Methods                  & Template-based & Energy-based & Our method \\ 
		\midrule
		Average inference time (s)    &  0.32       & $\sim$2400.00  & 0.69 \\			
		\bottomrule
	\end{tabular}
	\label{tab:test-time}
\end{table}

\section{Conclusion}
In this paper, we have proposed a novel attribute-conditioned layout GAN to synthesize graphic layouts by incorporating element attributes as conditions into both the generator and the discriminator. In addition, we introduce a novel element dropout strategy to enhance the capability of the discriminator in capturing local patterns, as well as several hand-crafted loss designs to facilitate layout optimization. We apply our model to generate attribute-conditioned layouts and adjust existing ones to new sizes. Experimental results well demonstrate the effectiveness of our method. 
In this work, we define the reading-order by simply following the left-to-right and top-to-down viewing order commonly used in banner ads. However, for designs containing many elements, defining the reading-order could be a complex problem. Future work include determining more content-adaptive reading-orders using theories such as visual attention and salience for automatic graphic design.

\bibliographystyle{IEEEtran}
\bibliography{egbib}

\end{document}